\documentclass[english]{article}

\usepackage{proceed2e}
\usepackage[margin=1in]{geometry}
\usepackage{times}
\usepackage{subcaption}

\usepackage{graphicx}
\usepackage{float, capt-of}
\graphicspath{ {./} }

\usepackage{mathtools}
\usepackage{algorithm}
\usepackage{algpseudocode}

\author{
    Sebastian Schulze \hspace{0.9em}  \, Owain Evans\\
University of Oxford\\
}

\DeclareMathOperator*{\argmax}{argmax} 
\newcommand{\comment}[1]{\ignorespaces}

\usepackage{natbib}
\bibpunct{(}{)}{;}{a}{,}{,}
\usepackage{hyperref}

\newcommand{\M}{\mathcal{M}}

\newcommand{\R}{\mathcal{R}}
\renewcommand{\P}{\mathcal{P}}
\newcommand{\thM}{\theta_{\M}}
\newcommand{\thP}{\theta_{\P}}

\newcommand{\sh}{\left\langle s,h \right\rangle}
\newcommand{\QM}{Q_{\M}}
\newcommand{\dmax}{d_{\mathrm{max}}}
\newcommand{\seq}{\,{=}\,}
\newcommand{\sgt}{\,{>}\,}
\newcommand{\shin}{\,{\in}\,}

\begin{document}
\title{Active Reinforcement Learning with Monte-Carlo Tree Search}
\maketitle
\begin{abstract}
Active Reinforcement Learning (ARL) is a twist on RL where the agent observes reward information only if it pays a cost. This subtle change makes exploration substantially more challenging. Powerful principles in RL like optimism, Thompson sampling, and random exploration do not help with ARL. We relate ARL in tabular environments to Bayes-Adaptive MDPs. We provide an ARL algorithm using Monte-Carlo Tree Search that is asymptotically Bayes optimal. Experimentally, this algorithm is near-optimal on small Bandit problems and MDPs. On larger MDPs it outperforms a Q-learner augmented with specialised heuristics for ARL. By analysing exploration behaviour in detail, we uncover obstacles to scaling up simulation-based algorithms for ARL.  
\comment{
For Arxiv: pdf images, acknowledgements, hyperref. Citations should be consistent -- e.g. only first initials and not first names.}
\end{abstract}

\section{INTRODUCTION}
\subsection{Motivation}
Imagine two treatments are being tested in a medical trial. The treatments are cheap but having doctors evaluate whether they worked costs \pounds 5,000 for each additional patient. Treatments are assigned using a Bandit design \citep{medicalBandits} and after 200 trials the difference in mean evaluation between the two treatments is tiny. Should the trial continue?

At some point an additional trial is not worth another \pounds 5,000. This cost of evaluating outcomes is not incorporated into standard Bandits. When playing Bandits, deciding whether to explore depends only on the estimated differences in expected (discounted) return between arms. The same is true for Reinforcement Learning in MDPs: the cost of providing a reward for a state-action pair is not a parameter of the learning problem.%\footnote{Incorporating such costs into Bandit-like problems is treated by \textit{partial monitoring}~\citep{piccolboni2001discrete}. We focus on adding  MDP case.}
This makes sense when the reward function is created all at once and \textit{offline}, as when it is hand-engineered. But if the rewards are created incrementally \textit{online}, as in the medical trial, then an important feature of the decision problem has been left out. 

\begin{figure}[t!]
\centering
\includegraphics[width=.85\columnwidth]{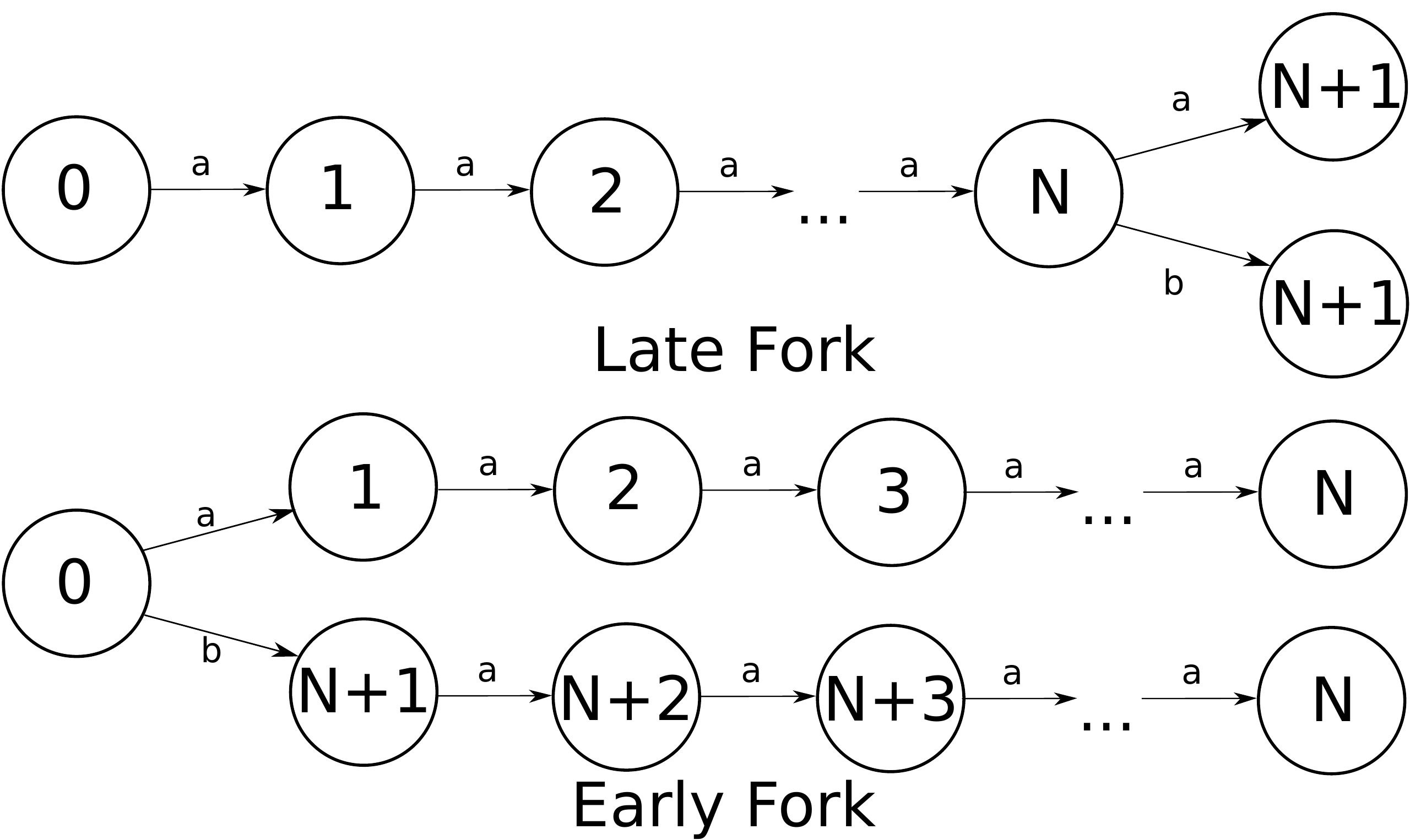}

\caption{Late Fork and Early Fork are deterministic, tabular MDPs. The edges indicate available actions. At only one state, the \textit{fork} (states $\mathrm{N}$ and 0) are two actions available.   The agent knows the transitions but not the rewards. In Late Fork, the agent should query only at the fork (all other actions are \textit{unavoidable}). In Early Fork, the agent should query everywhere, as all rewards contribute to Q-values at state 0.}
\label{fig:fork}
\end{figure}

Online construction of rewards is common in real-world Bandit problems: customers subjected to A-B testing may be paid to give feedback on new products \citep{scott2015multi}. Recent research, spurred by the difficulty of hand-engineering rewards, has formalised more general approaches to online reward construction. In \textit{Reward Learning}, a reward function is learned online from human evaluations of the agent's behaviour~\citep{DeepTAMER,HumanPref,saunders2017trial}. In \textit{Inverse Reinforcement Learning} (IRL) and \textit{Imitation Learning} \citep{apprentice,GAIL,evans2016learning}, humans provide demonstrations that are used to infer the reward function or optimal policy. These demonstrations can be provided offline or online
but the reward function is always specified \textit{incrementally}, as a set of human actions or trajectories.

In Reward Learning and IRL, the human labour required to construct rewards is a significant cost. How can this cost be reduced? Intuitively, if the RL agent can predict an action's reward then a human need not provide it. In \textit{Active Reinforcement Learning} (ARL), this choice of whether to pay for reward construction is given to the RL agent~\citep{activeRL}. It is the analogue of Active Learning, where an algorithm decides online whether to have the next data point labelled in a classification or regression task~\citep{activeLearning}. %An ARL agent should avoid paying for reward labels when they are unlikely to alter its policy.

%RL agents already have to trade off exploration and exploitation. ARL adds additional costs to exploration but one might wonder whether it changes the RL problem fundamentally. We will argue that different between ARL and RL is significant when it comes to practical learning algorithms. Various insights that fuel efficient RL algorithms do not carry over to ARL and so new ideas are needed to make ARL practical even in simple environments. 

\label{sec:arl-definition}
\subsection{ARL Definition and Illustration}

To fix intuition, we define the ARL problem here and elaborate on this definition in later sections. 
An active reinforcement learning (ARL) problem is a tuple $(S, A, \P, \R, \tau, c)$. The components $(S,A,\P,\R,\tau)$ define a regular Markov Decision Process (MDP), where $S$ is the state space, $A$ is the action space, $\P$ is the transition function, $\R$ is the reward function on state-action pairs, and $\tau$ is the time horizon. The component $c \sgt 0$ is a scalar constant, the ``query cost'', which specifies the cost of observing rewards. All components except $\P$ and $\R$ are initially known to the agent.  

ARL proceeds as follows. At time step $t$, the agent takes an action pair $(i_t,a_t)$, where $i_t \shin \{0,1\}$ and $a_t \shin A$, which determines a reward $r_t \sim \R(s_t, a_t)$ and next state $s_{t+1}\! \sim\! \P(s_t, a_t)$. If $i_t \seq 1$, the agent pays to observe the reward $r_t$, and so receives a total reward of $r_t - c$. If $i_t\seq 0$ the agent does not observe the reward; so if the agent did something bad it will not be \textit{knowingly} punished. The agent's total return after $T$ timesteps is defined as:
\vspace{-.5cm}
\[
\mathrm{Return}(T) := \sum\limits_{t=0}^{T}  \R(s_t,a_t) - i_{t}c .
\]
\vspace{-.5pt}
We emphasise that actions for which the agent did not observe the reward still count towards the return.

An ARL problem depends crucially on how query cost $c$ compares to the agent's expected total returns. When $c$ is large relative to the expected returns, the agent should never query and should rely on prior knowledge about $\R$. When $c$ is very small, the agent can use a regular RL algorithm and always query. In between these two extremes, the agent must carefully select a subset of actions to query and so RL algorithms are not readily applicable to ARL. Figure \ref{fig:fork} shows two MDPs (\textit{Early Fork} and \textit{Late Fork}) that illustrate the challenge of deciding which actions to query. RL algorithms perform sub-optimally on these MDPs unless effort is made to adapt them to ARL. 

This paper presents the following contributions:

\begin{enumerate}
\item
  We show that ARL for tabular MDPs can be reduced to planning in a Bayes-Adaptive MDP.
  
  \item  
  We adapt MCTS-based algorithm BAMCP~\citep{BAMCP} to provide an asymptotically optimal model-based algorithm for Bayesian ARL. 

\item
BAMCP fails in practice on small MDPs. We introduce BAMCP++, which uses smarter model-free rollouts and substantially outperforms BAMCP.  

\item
We benchmark BAMCP++ against model-free algorithms with ARL-specific exploration heuristics. BAMCP++ outperforms model-free methods on random MDPs. 

\end{enumerate}

\subsection{Related Work}
How does ARL (as defined above) related to regular RL? In regular RL there is no cost for deciding to observe a reward. Yet regular RL does involve ``active learning'' in the more general sense: the agent decides which actions to explore instead of passively receiving them. So techniques for exploration in regular RL might carry over to ARL. 

Unfortunately, most practical algorithms for regular RL use heuristics for exploration such as $\epsilon$-greedy, optimism~\citep{UCB1,BEB}, and Thompson sampling~\citep{PSRL}. While these heuristics achieve near-optimal exploration for certain classes of RL problem~\citep{bubeck2012regret, azar2017minimax}, they are not directly applicable to ARL, as explained in Section~\ref{sec:algorithms}. There are RL algorithms that try to explore in ways closer to the decision-theoretic optimum. Various algorithms use an approximation to the Bayesian value of information~\citep{srinivas2009gaussian, BayesianQ} and so relate to our Section~\ref{sec:algorithms}. An alternative non-Bayesian approach is to have the agent learn about the transitions to which the optimal policy is most sensitive~\citep{epshteyn2008active}.

There is a substantial literature on active learning of rewards provided online by humans~\citep{wirth2017survey,dragan2017robot}.~\cite{ActiveRewards} learn a reward function on \textit{trajectories} (not actions) from human feedback and use Bayesian optimization techniques to select which trajectories to have labelled.~\cite{dorsa2017active} learn a reward function on state-action pairs and their agent optimizes actions to be informative about this function. These reward-learning techniques are aimed at continuous-state environments and do not straightforwardly transfer to our tabular ARL setting. Our work also differs from~\citeauthor{dorsa2017active} in that we optimize for informativeness about the optimal policy and not the true reward function. As Figure~\ref{fig:fork} illustrates, if some states are unavoidable then their reward is irrelevant to the optimal policy. 

There is also work applying active learning to tabular RL with human teachers but where human input is quite different than in the ARL model~\citep{subramanian2016exploration,judah2012active}.

\label{sec:background}
\section{BACKGROUND}

This section reviews Bayesian RL and the BAMCP algorithm. Later we cast ARL as a special kind of Bayesian RL problem and apply BAMCP to ARL.

\label{sec:BAMDP}
\subsection{Bayesian RL}
An MDP is specified by $\M \seq (S,A,\P,\R,\tau)$, with components defined in Section \ref{sec:arl-definition}. While our algorithms apply more generally, this paper focuses on finite, episodic MDPs \citep{PSRL}, where $\tau$ is the episode length. A Bayesian RL problem \citep{BRLsurvey,BAMCPthesis} is specified by an MDP $\M$ and an agent's prior distribution $b_0(\thP)$ on the transition function parameters $\thP$. The agent's posterior at timestep $t$ is then given by $b_t(\thP) \seq b_t(\thP | h_t) \propto \mathcal{L}(h_t | \thP )b_0( \thP )$, where $\mathcal{L}(h_t | \thP)$ is the likelihood of history $h_t$ given the transition function with parameters $\thP$. 

The Bayesian RL problem can be transformed into an MDP planning problem by augmenting the state space with the agent's belief and the transition function with the agent's belief update. The resulting MDP is defined by $\M^{+} \seq \big \langle S^+, A, \P^+, \R, \tau \big \rangle$ and is called a \textit{Bayes-Adaptive MDP }(BAMDP), where:

\begin{itemize}
\item $S^+$ is the set of hyperstates $S \times \thP$;
\item $P^+$ is the combined transition function between states and beliefs: $S^+\,\times\,A\,\times\,\R\,\times\,S^+\,\rightarrow\,[0,1]$; and
\item The initial hyperstate is determined by the initial distribution over $S$ and the prior $b_0$ on the transition function.
\end{itemize}

\subsection{BAMCP: MCTS for Bayesian RL}

BAMCP is a Monte Carlo Tree Search (MCTS) algorithm for Bayesian RL~\citep{BAMCP}. It converges in probability to the optimal Bayesian policy (i.e.\ the optimal policy for the corresponding BAMDP) in the limit of infinitely many MC simulations. In experiments, it has achieved near state-of-the-art performance in a range of environments \citep{BenchBRL,BAMCPthesis}.   

\begin{figure*}
\centering
\includegraphics[width=16cm]{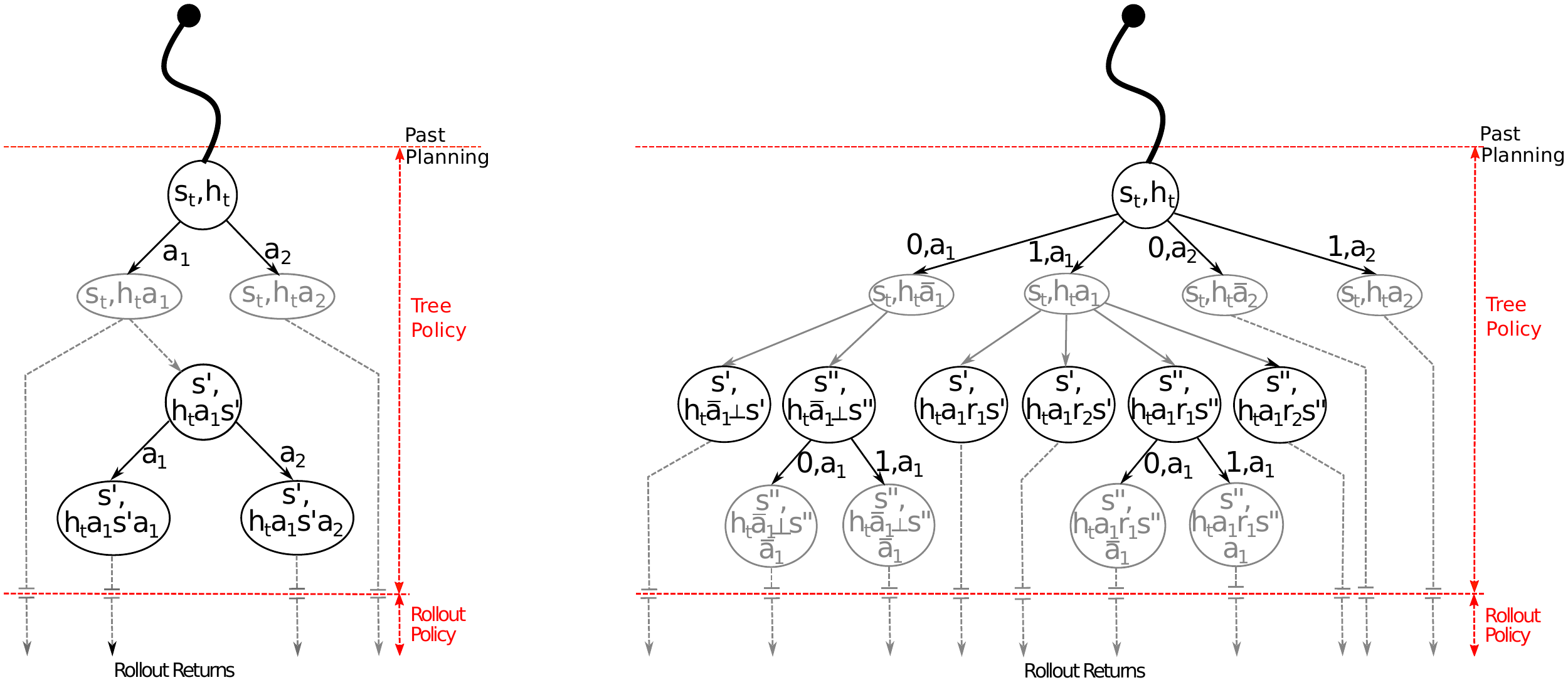}
\caption{Search tree for BAMCP applied to regular RL (\textbf{Left}) and applied to ARL (\textbf{Right}). Nodes in black correspond to a history $h_t$ and current state $s_t$. In the ARL tree, query actions have form $(1,a_t)$ and cause extra branching because they result in multiple possible observed rewards as well as multiple possible state transitions. At leaf nodes, actions are selected by a rollout policy.}
\label{fig:mcts}
\end{figure*}

At any given timestep BAMCP attempts to compute the Bayes-optimal action for the current state under the agent's posterior $b_t$. As is common for work on Bayesian RL, this posterior is only over the transition function and not the reward function.\footnote{For our experiments in ARL the agent will always be uncertain about the reward function.} BAMCP is an online algorithm. At each timestep, it updates the posterior on an observation from the real MDP and then uses MCTS to simulate possible futures using models sampled from this posterior. The MCTS builds a search tree mapping histories to value-function estimates (see Fig~\ref{fig:mcts}). A node corresponds to a posterior belief $b_t$ and current state action $(s_t,a_t)$, and for each node the algorithm maintains a value estimate $Q(\left\lbrace s_t, b_t \right\rbrace, a_t)$ and visit count $N(\left\lbrace s_t, b_t \right\rbrace, a_t)$. BAMCP's behaviour can be specified in four steps:
\vspace{-5pt}
\begin{enumerate}
\item Node selection: At any node BAMCP chooses to expand the subtree for the action chosen by a UCB policy. In particular, when at node $n \seq \left\lbrace s_t, b_t \right\rbrace$, the algorithm expands the action given by:
\vspace{-3pt}
\[
  \argmax_a Q(n,a) + u \sqrt{\frac{\log(\sum_a N(n, a))}{N(n,a)}}
 \]
\vspace{-3pt}
\noindent where $u$ is an exploration constant. 

\item Expansion: This node selection continues until it reaches the final timestep of the episode or a leaf node. At leaf nodes exactly one child node is added per simulation.

\item Rollouts: If additional steps outside the tree need to be simulated, a rollout policy, trained by running Q-learning on observations from the real MDP, selects actions. No new nodes are added during the rollout phase. 

\item Backup: After the rollout, value estimates of tree nodes along the trajectory are updated with the sampled returns. A simple average over all trajectories is computed. 
\end{enumerate}
BAMCP also uses {\em root sampling} and {\em lazy sampling} to improve efficiency~\citep{BAMCP}. 

\section{ALGORITHMS FOR ARL}
\label{sec:algorithms}

\subsection{Reducing ARL Problems to BAMDPs}

We consider Active RL (defined in Section \ref{sec:arl-definition}) in the Bayesian setting, where the agent has a prior distribution over the reward and transition functions. This is similar to a Bayesian RL problem. Actions in ARL reduce to regular RL actions by crossing each regular action with an indicator variable. But unlike in RL, an ARL agent does not always observe a scalar reward. To accommodate this, we introduce the null reward ``$\bot$''. If the agent takes an action without querying, it receives a reward $\bot$. The definition of the agent's belief update is modified to not update on $\bot$. With this minor emendation, Bayesian ARL can be reduced to an MDP in an augmented state-space exactly as in Section \ref{sec:BAMDP}.

\subsection{RL Algorithms Fail at ARL}
Can we apply Bayesian RL algorithms to Bayesian ARL? Many such algorithms can be straightforwardly adapted to deal with the null reward and produce well-typed output for ARL. Yet naive adaptations often fail pathologically. For instance, they might never choose to query and hence learn nothing. Here are some principles used in RL algorithms that lead to pathologies in ARL.

 \noindent \textbf{Optimism in the face of uncertainty} \newline 
  Optimism means adding bonuses to more uncertain rewards and taking optimal actions in the resulting optimistic MDP \citep{BEB,BOLT,UCB1}. An optimal agent in a \textit{known} MDP never queries. Optimism treats the optimistic MDP as (temporarily) known and hence optimism applied to ARL never queries.

  \noindent \textbf{Thompson Sampling (PSRL)} \newline 
Thompson Sampling samples from the posterior on MDPs and plans in the sampled MDP \citep{PSRL,strens}. This fails for the same reason as optimism.

 \noindent \textbf{Model-free TD-learning with random exploration} \newline 
TD-learning is described in \cite{suttonBarto}. The $Q^*$-value of querying an action is always lower than the value of not querying the same action. So for every action, a TD-learner learns to avoid querying the action and so fails when some actions must be queried many times. 

% \noindent \textbf{Myopic Value-of-Information} \newline 
% Value-of-information means estimating the Bayesian expected return of exploratory actions by explicitly modelling their information gain~\citep{BayesianQ,BRLsurvey}. Myopic (one-step) approaches assume the agent gains information only from the current action and thereafter exploits. This sometimes leads to highly sub-optimal behaviour in ARL because it stops exploring when the agent can not learn enough from a single query to change their policy. 

\subsection{Applying BAMCP to ARL}
BAMCP is simple to adapt to Bayesian ARL and does not lead to obvious pathologies like the principles above. In fact, it converges in the limit to the optimal Bayesian policy for the Bayes-Adaptive MDP derived from the ARL problem. Adapting BAMCP to ARL requires a few modifications of \cite{BAMCPthesis} which are depicted in Figure~\ref{fig:mcts}. First, we explicitly model uncertainty over both the reward function $\R$ and transition function $\P$. Second, the rollout policy only considers non-querying actions (as querying is pointless for rollouts that do not learn). Third, querying is incorporated into Monte-Carlo simulations. When simulating a trajectory, each action $a \shin A$ can be queried or not queried, as represented by indicator $i$. If the action is not queried, the search tree may not branch (since there is no reward observation) but the reward backup is still performed. If the action is queried, its reward is observed and reduced by the query cost $c$.
\begin{algorithm}[H]
\begin{algorithmic}
\Procedure{BAMCP-PP}{$T$}
\State $h\leftarrow \left\lbrace \right\rbrace$, $t \leftarrow 0$, $s \leftarrow s_0 $
\State Initialise $\QM$ randomly.
\Repeat
\State $i,a \leftarrow$ MCTS-SEARCH$(\sh)$
\State $r,s' \sim  P(\cdot | s,i,a)$
\State Append $(s,i,a,r,s')$ to $h$
\State $t \leftarrow t + 1$
\If {$i = 1$} 
\State Q-LEARN-UPDATE($\QM, s, a, r$)
\EndIf
\Until{$t = T$}
\EndProcedure
\State
\Procedure{MCTS-Search}{$\sh$}
\Repeat
 \State $\thM \sim b(\thM | h)$
 \State $Q_\pi \leftarrow \QM$
 \State SIM$(\sh,\thM, 0)$
\Until{Time-out}
\State \textbf{return} $\argmax\limits_{i,a} Q(\sh, i, a)$ 
\EndProcedure
\State
\Procedure{Sim}{$\sh$, $\thM$, $d$}
\If{$d> \dmax $}
\State \textbf{return} 0
\EndIf
\If{$N(\sh)$ = 0}
\State $0$, $a \leftarrow \pi_{ro}(\sh)$
\State $r, s' \leftarrow P(\cdot|s,a,\thM)$
\State $R \leftarrow r$ + ROLLOUT($\left\langle s',ha\perp s' \right\rangle$,
$\thM$,$d+1$)
\State Update$(\sh, R)$
\State \textbf{return} R
\EndIf
\State $i,a \leftarrow \argmax\limits_{i,a} Q(\left\langle s, h \right\rangle,i,a) + u \sqrt{\frac{\log(N(\left\langle s, h \right\rangle))}{N(\left\langle s, h \right\rangle,i,a)}}$
\State $r, s' \leftarrow P(\cdot|s,a,\thM)$
\If{$i$ = 0}
\State $R \leftarrow r + $SIM$(\left\langle s',ha\perp s' \right\rangle$,$\thM$,$d+1)$
\Else
\State $R \leftarrow r + $SIM$(\left\langle s',hars' \right\rangle$,$\thM$,$d+1) - c$
\State Q-LEARN-UPDATE($Q_\pi, s, a, r$)
\EndIf
\State  Update$(\sh, i,a, R)$
\State \textbf{return} $R$
\EndProcedure

\end{algorithmic}
\caption{BAMCP++ Algorithm\newline Main procedure BAMCP-PP is applied for $T$ timesteps to unknown MDP $\M$. Posterior over $\M$ is represented by $b(\thM|h)$. The procedure Q-LEARN-UPDATE is the standard Q-learning update.}\label{alg:bamcppp}
\end{algorithm}

\setcounter{algorithm}{0}
\begin{algorithm}[H]
\floatname{algorithm}{Algorithm}
\begin{algorithmic}
\Procedure{Rollout}{$\sh, \thM, d$}
\If{$d > \dmax$}
\State \textbf{return} 0
\EndIf
\State $0,a \leftarrow \pi_{ro}(\sh)$
\State $r, s' \leftarrow P(\cdot|s,a,\thM)$
\State $R \leftarrow r + \mathrm{ROLLOUT}(\left\langle s',ha\perp s' \right\rangle,\thM, d+1)$
\State Update$(\sh, R)$
\State \textbf{return} $R$
\EndProcedure
\State
\Function{$\pi_{ro}$}{$\sh$}
\State \textbf{return} $a \sim \mathrm{SoftMax}(Q_\pi(s,\cdot))$ 
\EndFunction
\end{algorithmic}
\caption{BAMCP++ cont'd}\label{bamcp}
\end{algorithm}

\subsection{Algorithm for ARL: BAMCP++}

As we show in Section \ref{sec:experiments}, BAMCP performs poorly on ARL. We introduce \textit{BAMCP++}~(Algorithm \ref{alg:bamcppp}), which builds on BAMCP and leads to improved estimates of the value of querying actions.
The first new feature of BAMCP++ is {\em Delayed Tree Expansion}. UCB tree expansion often avoids query actions, because it is hard to recognise their value when estimating via noisy rollouts. To address this, we accumulate the results of multiple rollouts from a leaf node before letting UCB expand the actions from that node. This reduces the variance of value estimates, helping to prevent query actions from being prematurely dismissed. The second new feature of BAMCP++ addresses a problem with the rollouts themselves. 

\subsubsection{Episodic Rollouts}
BAMCP's rollout policy is responsible for value estimation in parts of the state space not yet covered by the MCTS search tree. Returns from a rollout are used to initialise leaf nodes and are also propagated back up the tree.

BAMCP's rollout policy consists of a Q-learner trained on observations from the real MDP. This can result in a \textit{vicious circle} when applied to ARL: (i) the Q-learner can only learn from the real MDP if the agent chooses to query; (ii) the agent only chooses to query if simulated queries lead to higher reward; (iii) simulated queries only lead to higher reward if the information gained is exploited and random rollouts do not exploit it. Our experiments suggest this happens in practice: BAMCP queries far too little. Related to the vicious circle, BAMCP's rollouts do not share information across related hyperstates. After getting a big reward ten out of ten times from one simulated Bandit arm, the rollout is just as likely to choose the other arm. 

In Episodic Rollouts, the rollout policy is still a Q-learner. But instead of just training on the real MDP, we also train on the observations from the current MC simulation. Let $\QM$ denote a Q-learner trained on the real MDP up to timestep $t$. For each MC simulation, the rollout is performed by a distinct Q-learner $Q_{\pi}$ that is initialised to $\QM$ but then trained by Q-learning on observations in the simulated\footnote{This is the \textit{root-sampled} MDP.} MDP $\M_{\mathrm{sim}}$ (see Q-LEARN-UPDATE applied to $Q_{\pi}$ in Algorithm~\ref{alg:bamcppp}). This simulation consists of repeated episodes of $\M_{\mathrm{sim}}$  and so $Q_{\pi}$ gradually learns a better policy for $\M_{\mathrm{sim}}$, sharing information across hyperstates and exploiting querying actions. The rollout's actions are sampled from a Boltzmann distribution.\footnote{Since the search tree eventually covers the entire state space (due to UCB), we can freely modify the rollout policy can without removing the asymptotic guarantees of MCTS.}

Episodic Rollouts use a model-free agent that learns during simulation, at the cost of a slower rollout. Having a fast model-free agent to guide model-based simulations is also central to AlphaZero~\citep{alphago,alphazero,slowandfast}, where the model-free network is trained to predict the result of MCTS simulations. 

\subsection{Model-free Agents for ARL}

%based on TD-learning and random exploration play a central role in RL, achieving state-of-the art performance in complex environments \citep{IMPALA,DOTA,mujoco}\comment{appropriate paper?}. We want to investigate whether model-free agents can achieve reasonable performance in tabular ARL.

As noted above, model-free agents such as $\epsilon$-greedy Q-learners can fail pathologically at ARL. We want to investigate whether Q-learners augmented with querying heuristics can perform well on ARL. 

The {\em First-N Heuristic} queries each state-action pair on the first $N$ visits. The hyperparameter $N$ can be tuned empirically or set using prior knowledge of the transition function and the variance of reward distributions.

The {\em Mind-Changing Cost Heuristic} (MCCH) of~\cite{activeRL} is based on bounding the value of querying and is closely related to the Value of Information heuristic \citep{BayesianQ}. After enough timesteps, an optimal Bayesian ARL agent may stop querying because the value of information (which decreases over time) does not exceed the query cost (which is constant). Likewise, MCCH computes an approximate upper bound on the value of querying and avoids querying if the bound exceeds the query cost. The bound is based on the number of episodes remaining $E$, the value $Q_{\mathrm{max}}$ of the best possible policy (consistent with existing evidence), the value $\bar{V_{t}}$ of the currently known best policy, and finally the number of queries $m$ required for the agent to learn they should switch to $Q_{\mathrm{max}}$.
%In Bandits, the quantities $Q_{\mathrm{max}}$, $Q{_\mathrm{cur}}$ and $m$ can be computed fairly accurately and MCCH is a good heuristic~\citep{activeRL}. In the MDP case, it is much harder to approximate these quantities. 
The quantity $Q_{\mathrm{max}}$ can be upper-bounded by the total reward possible in an episode (given the maximum reward $R_{\mathrm{max}}$). Since $m$ is difficult to approximate without prior knowledge, we replace it with a hyperparameter $\mu \sgt 0$ that needs to be tuned. If the agent follows MCCH for MDPs, it queries whenever:
\[
c \mu < E( Q_{\mathrm{max}} - \bar{V_{t}} )
\]
The First-N Heuristic and MCCH can be combined with any model-free learner. In our experiments, we use an $\epsilon$-greedy Q-learner. For First-N, if a state-action has been queried $N$ times, it cannot be chosen for exploratory actions. For MCCH, the agent follows $\epsilon$-greedy up until it stops querying at which point it just exploits using its fixed Q-values. 

\section{EXPERIMENTS}
\label{sec:experiments}

We test BAMCP and BAMCP++ on Bandits and then investigate the scalability of BAMCP++ on a range of larger MDPs.  

\subsection{BAMCP vs.\ BAMCP++ in Bandits}
In the ARL version of multi-armed Bandits, the agent decides both which arm to pull and whether to pay a cost to query that arm. Optimal behaviour in ARL Bandits has a simple form: the agent queries every action up to some point and thereafter never queries~\citep{activeRL}. We test BAMCP against BAMCP++ on a two-arm Bernoulli Bandit, with parameters for the two arms $p\seq\{0.2,0.8\}$ and a query cost $c\seq0.5$. The total number of trials (which is known) varies up to 40. Both algorithms have $\mathrm{Beta}(0.5,0.5)$ priors over arm parameters and use 200,000 Monte-Carlo simulations. Gridsearch was used to set the UCB hyperparameter $u$ and BAMCP++'s delayed tree-expansion parameter.

\subsubsection{BAMCP++ is near optimal}
\begin{figure}[t!]
\centering
\includegraphics[width=7cm]{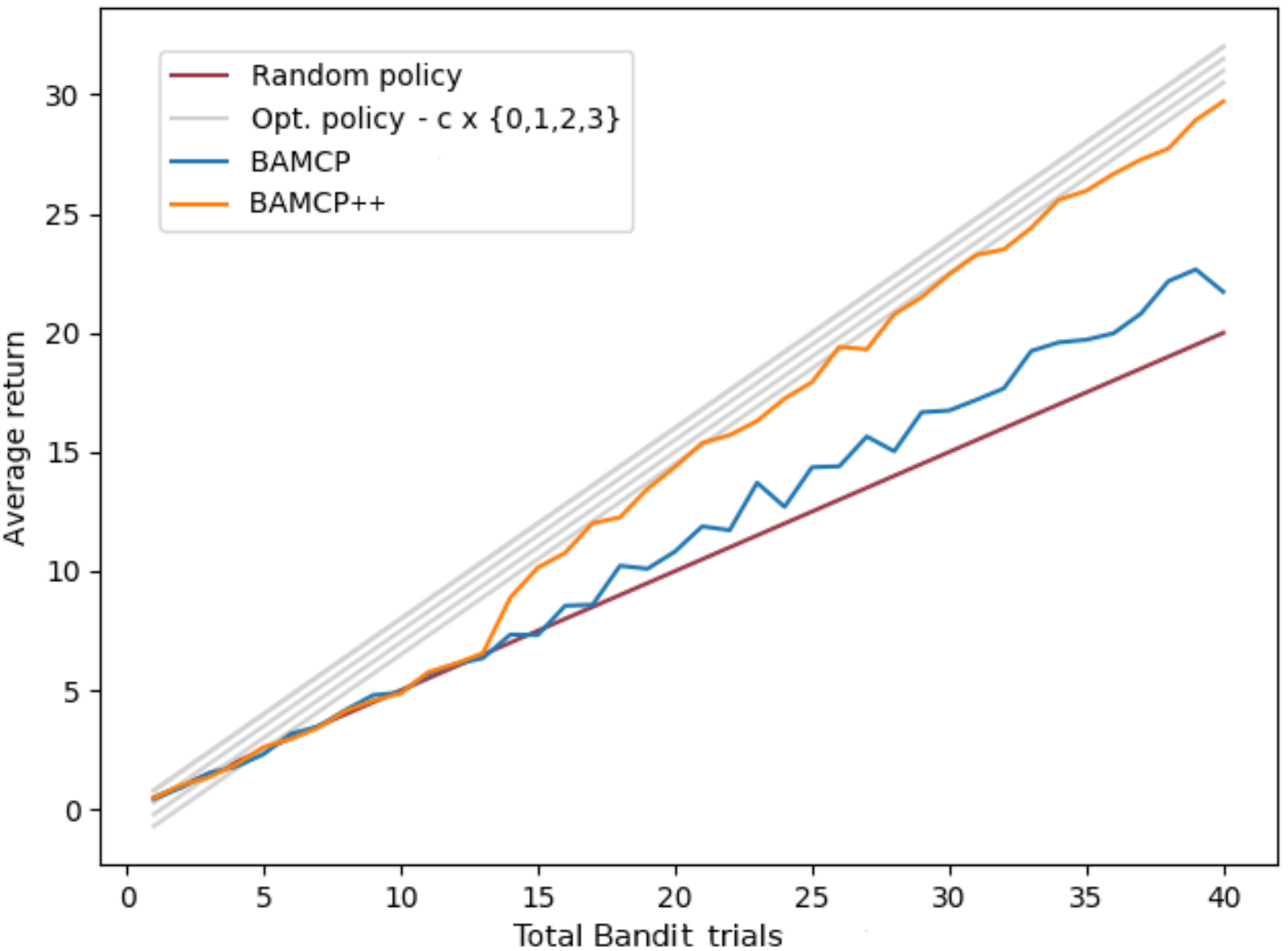}
\caption{Mean returns over 100 runs on 2-arm Bernoulli ARL Bandit with $p \seq \{0.2, 0.8\}$, $c \seq 0.5$, and with varying horizon (total trials). 
%Grey lines are optimal policy if it had made up to three queries.
}
\label{fig:bandit-return}
\end{figure}

Figure \ref{fig:bandit-return} shows returns averaged over 100 repeats of the same ARL Bandit (for horizons up to 40). We compare BAMCP and BAMCP++ to the optimal policy (which always pulls the best arm and never queries) and to the optimal policy minus the cost of up to three queries (for a fairer comparison). The optimal policy is distinct from the Bayes optimal policy, which is the ideal comparison but is hard to compute~\citep{activeRL}. BAMCP++ is mostly close to the optimal policy minus three queries, whereas BAMCP is closer to the random policy. 

\begin{figure}[t!]
\centering
\includegraphics[width=8cm]{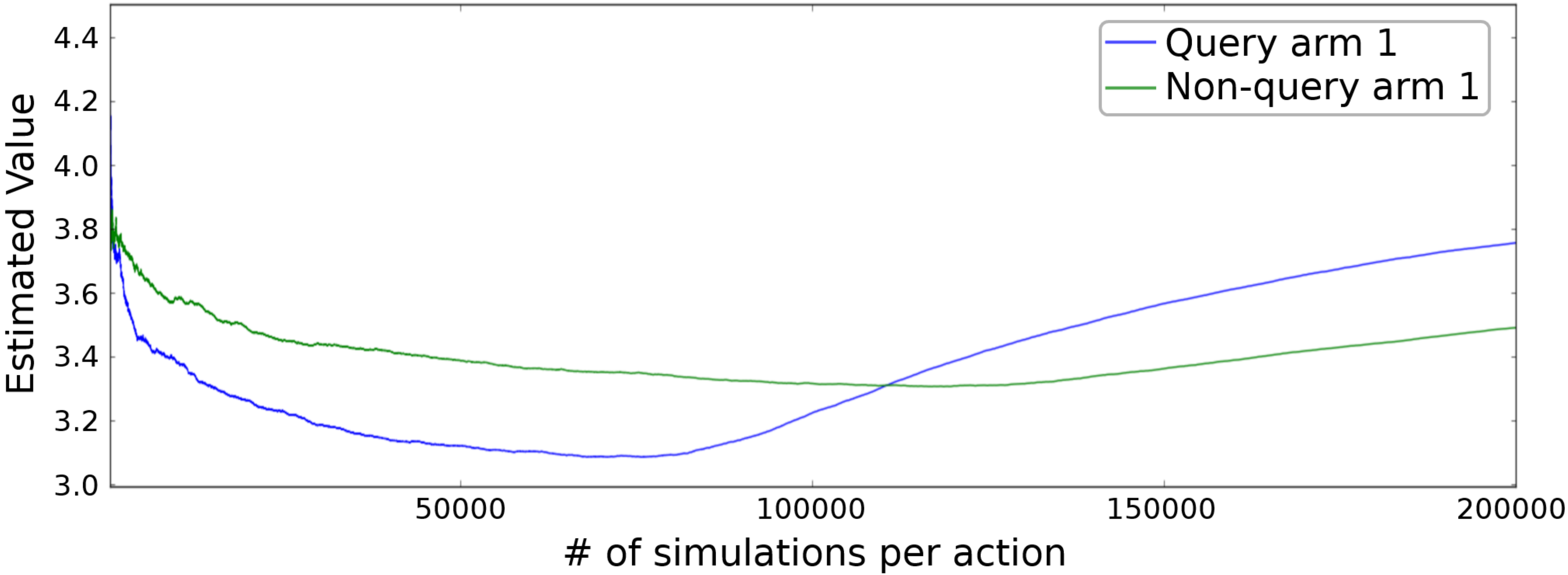}
\caption{Estimated Q-value for query/non-query on first trial of Bandit after a given number of MCTS simulations. Horizon is set to $T \seq 15$. %MCTS incorrectly estimates non-querying to be superior up to 100,000 simulations because querying is worse unless the agent exploits flawlessly.
}
\label{fig:q-values-time}
\end{figure}

While BAMCP++ is near-optimal for horizon $T \sgt 15$, it is suboptimal for smaller horizons. What explains this? For sufficiently small $T$, the Bayes optimal agent does not query and performs randomly. However, for $T\seq 12$ the Bayes optimal agent would query and so BAMCP++ falters. The difficulty is that querying is only optimal if the agent performs flawlessly after the query. Hence many MCTS samples are needed to recognise that querying is Bayes optimal (as most trajectories that start with querying are bad). This is illustrated in Figure \ref{fig:q-values-time}, which shows the estimated BAMDP Q-values for query and non-query actions in the first timestep for $T\seq 15$. Even after 100,000 simulations, non-querying is (incorrectly) estimated to be superior.

\begin{figure}[t!]
\centering
\includegraphics[width=7cm]{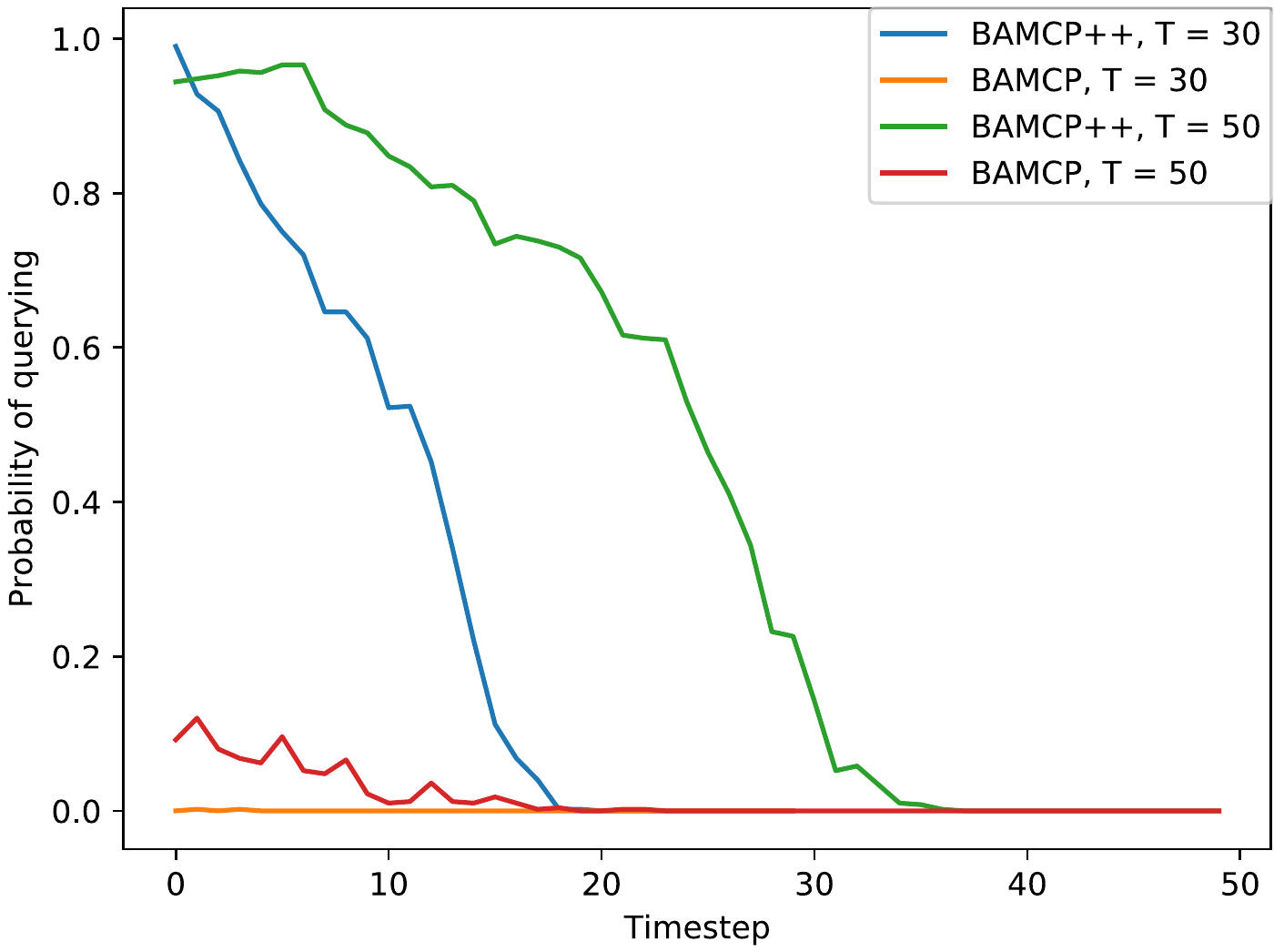}
\caption{Probability of querying at each timestep in ARL Bandit (see Fig~\ref{fig:bandit-return}) with fixed horizon $T \seq 30$ and $T \seq 50$. 
%Optimal policy queries with prob 1 for first few timesteps and stops querying near horizon (near $t$=30 and $t$=50).
} 
\label{fig:bandit-query}
\end{figure}

BAMCP is outperformed by BAMCP++. Figure \ref{fig:bandit-query} shows the probability of queries for $T \seq 30$ and $T \seq 50$. For these horizons, the Bayes optimal agent queries at the first few timesteps with probability one. Yet BAMCP almost never queries ($T \seq 30$) or queries with low probability ($T \seq 50$). BAMCP (unlike the random agent) exploits information gained from queries but because it fails to recognise the value of queries it never gains much information.

\subsubsection{BAMCP's problems in regular RL}
\begin{figure}[t!]
\centering
\includegraphics[width=6.5cm]{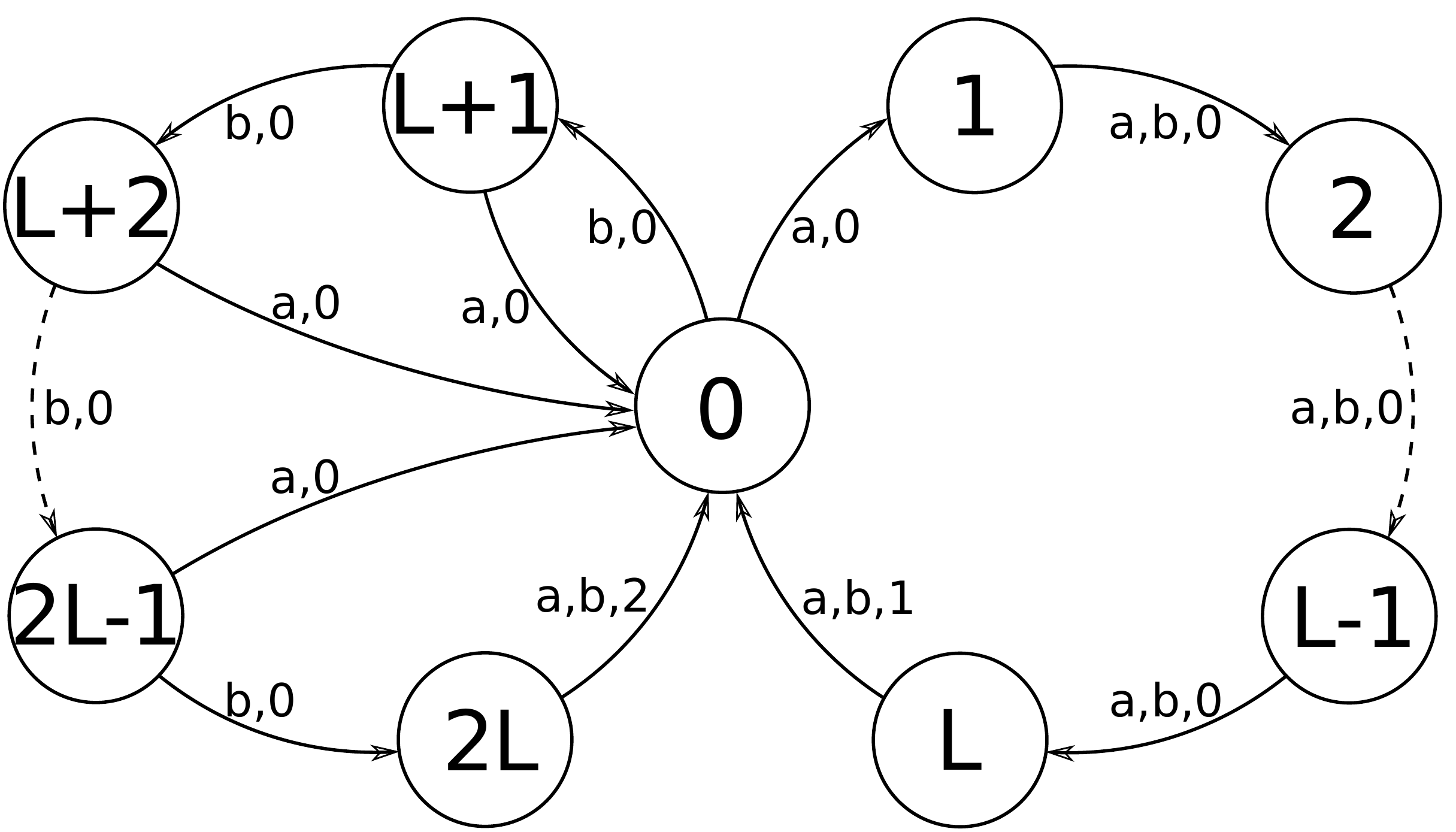}
\caption{The Double-Loop MDP~\citep{BayesianQ} for RL consists of two loops of length $L$. The optimal policy traverses the entire left loop.}
\label{fig:double-loop}
\end{figure}

Is the failure of BAMCP in Bandits due to a special feature of ARL, or does BAMCP fail at related problems in regular RL? We tested BAMCP on the Double-Loop (Fig~\ref{fig:double-loop}), an RL environment that poses a similar challenge to ARL Bandits. For this environment the agent knows the rewards and has a Dirichlet prior on the transition probabilities. While BAMCP achieved excellent performance on Double-Loops with $L\seq4$~\citep{BAMCPthesis,BenchBRL}, we test it up to $L\seq10$. We set the UCB parameter $u\seq3$ and the number of MC simulations to 10,000 (following \citeauthor{BAMCPthesis}). Figure \ref{fig:double-loop} shows that BAMCP's performance drops rapidly after $L\seq4$ and ends up no better than a simple model-free Q-learner. How is this poor performance related to ARL? Suppose the agent believes (after trying both loops) that reaching state $2L$ is likely worse than the right loop. The reason to explore $2L$ is that if it is better it can be exploited many times. But unless MCTS simulates that systematic exploitation the agent will not explore.
\comment{is that explanation right?}
\comment{Not quite: In the double loop environment the agent needs to repeatedly decide to explore the second loop. In order for this to happen MCTS needs to run rollouts in which it recognises the potential for higher reward at the end of that loop (despite potential information that early parts of the loop have 0 rewards) as well as the following exploitation. The same sort of goal directed simulation is necessary for the recognition of query value - not only does the agent need to simulate the query in the first place, but the simulation needs to then exploit the gathered knowledge}

%\comment{What's time horizon in our double loop experiments? What's horizon in Guez experiments? Also, is there loop diagram from Guez? If so, we should credit them.}
%\comment{The horizon varies with the loop length but it is a fixed number of episodes... they run 1000 steps (200 episodes) and I think we have the same or something similarly large, there is a double picture which is for fixed loop size and attributed to dearden 1998}

%\comment{TODO: note somewhere the result that UCT can fail to find a good path hidden in a low reward region.Coquelin and R. Munos. Bandit algorithms for tree search.}


\subsection{Benchmarking BAMCP++ and Model-Free Algorithms}

Having shown that BAMCP++ does well on ARL Bandits, we test it on more complex MDPs with unknown transition dynamics and compare it against model-free algorithms.

\subsubsection{BAMCP++ on Late Fork}
\begin{figure}[t!]
\centering
\includegraphics[width=8cm]{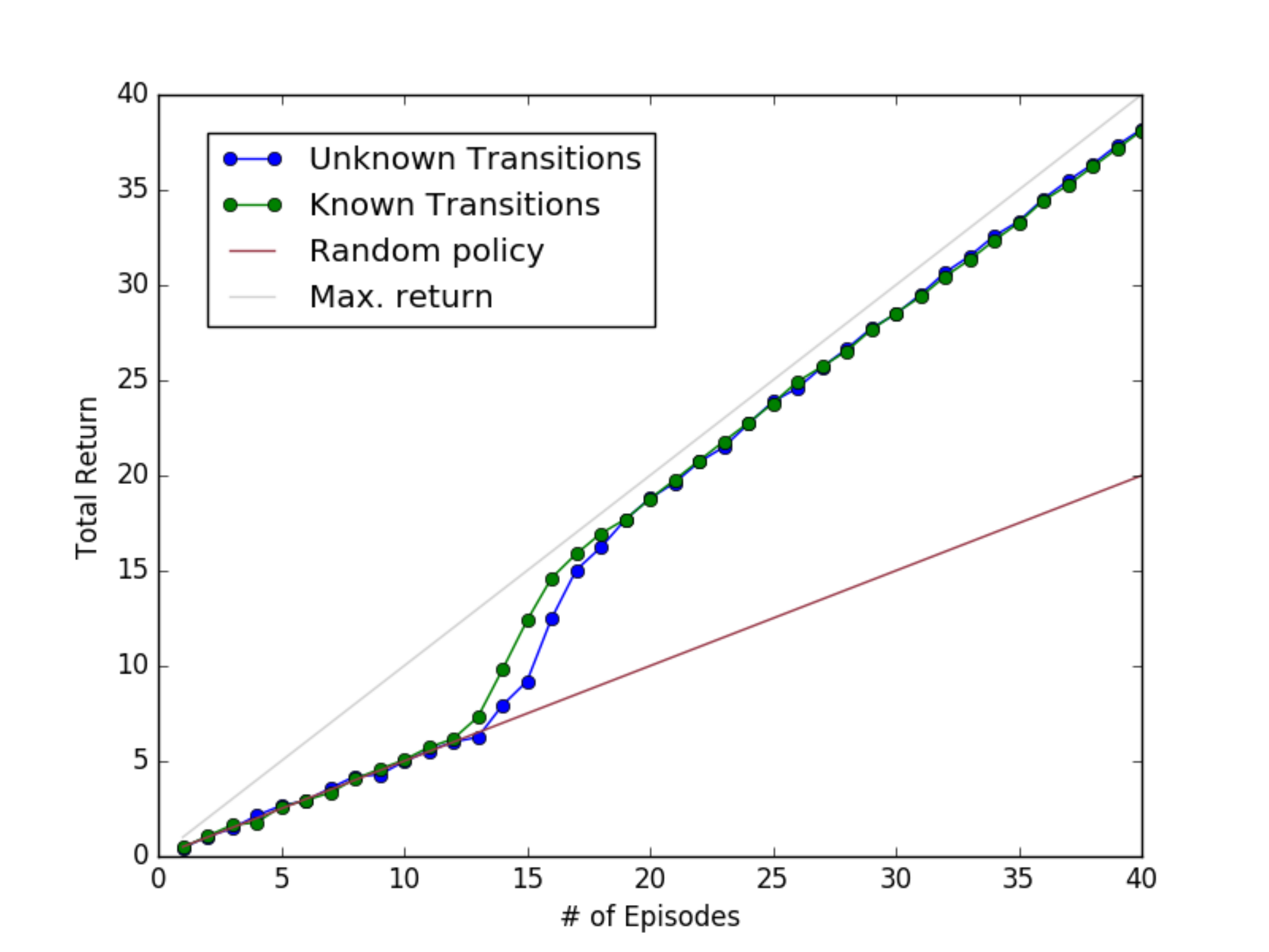}
\caption{Average returns of BAMCP++ on Late Fork ($N \seq 2$) compared to the optimal policy (``Max return'') as function of horizon $T$.}
\label{fig:late-fork-reward}
\end{figure}

We test BAMCP++ on Late Fork (Fig~\ref{fig:fork}) with $N\seq2$. This is a 3-state MDP, where the first two actions are unavoidable and should not be queried. The query cost is $c\seq0.5$. In the condition ``Known Transitions'', all transitions are known and only the rewards for each action are unknown. In ``Unknown Transitions'', the agent knows which actions are available at each state but not where the actions lead. The priors are $\mathrm{Beta}(0.5,0.5)$ for rewards and symmetric Dirichlet with parameter $\alpha \seq 0.5$ for transitions. Figure \ref{fig:late-fork-reward} shows total returns averaged over 50 runs for different horizons. (The number of episodes plays the same role as the number of trials in Bandits).  

\begin{figure}[t!]
\centering
\includegraphics[width=65mm]{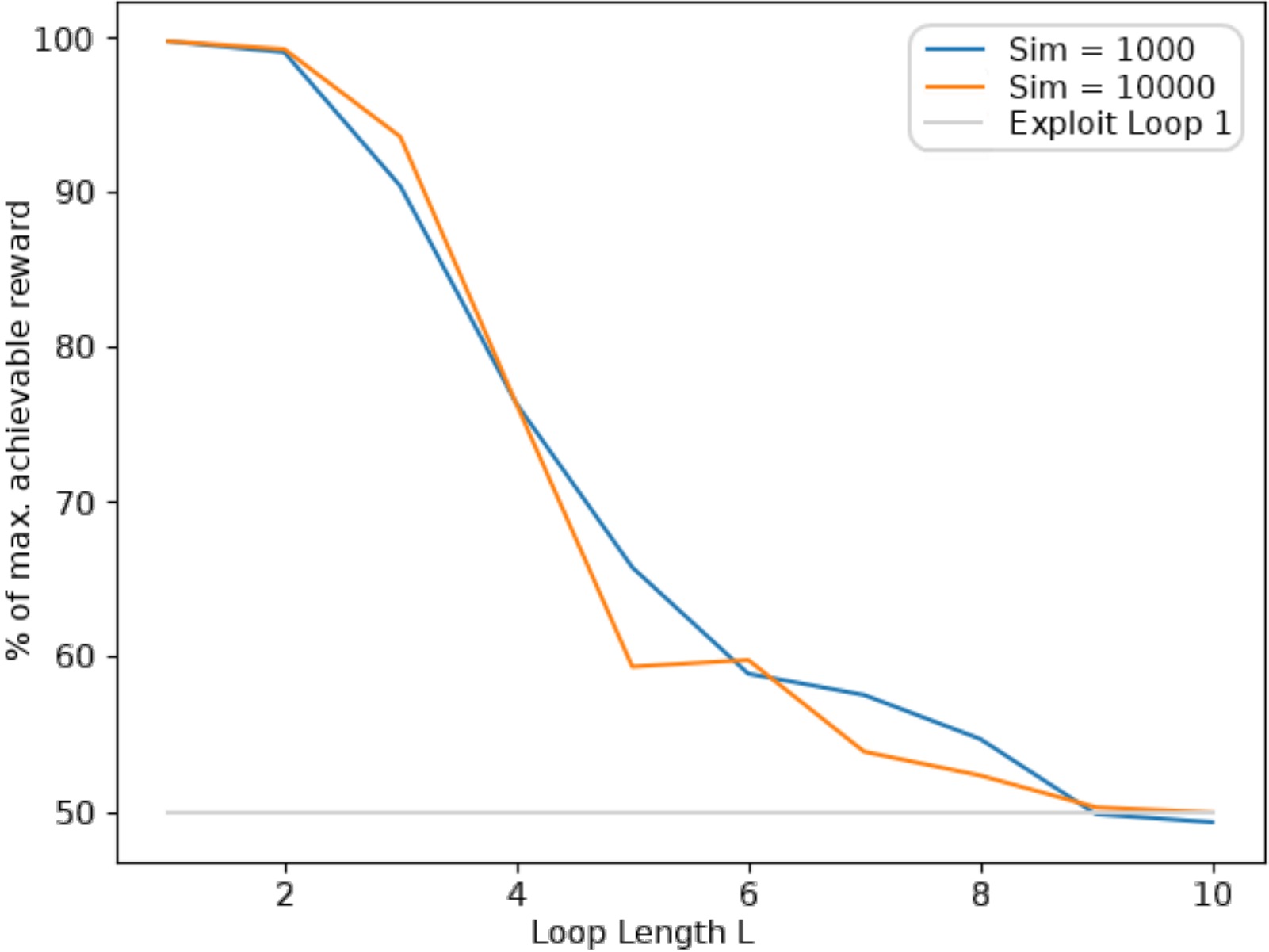}
\caption{Average performance of BAMCP (regular RL) on the Double-Loop with different simulation budgets and varying loop length $L$. Grey lines represents expected score of a Q-learner.}
\label{fig:double-loop-results}
\end{figure}

\begin{figure}[t!]
\centering
\includegraphics[width=6cm]{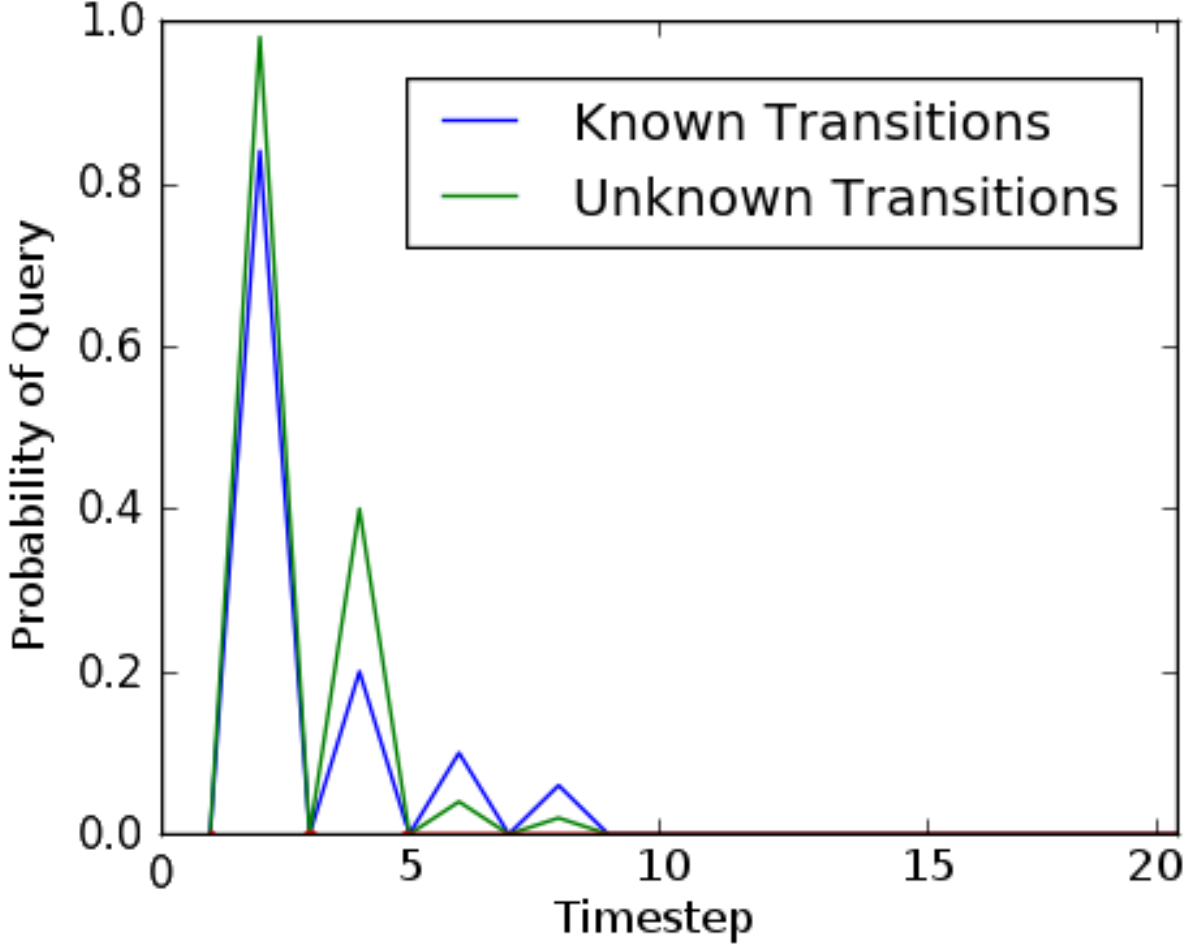}
\caption{Probability of query actions for BAMCP++ on each timestep of Late Fork (with horizon $T \seq 20$).
% TODO this graph needs to be under or next to Y_2_time
}
\label{fig:prior-queries}
\end{figure}

BAMCP++ achieves close to the optimal policy when the horizon $T$ is above 17. But does it explore in the Bayes optimal way? Figure \ref{fig:prior-queries} shows the probability of querying actions at each timestep in a setting with horizon $T \seq 20$ episodes, which corresponds to the mid-point on the x-axis of Fig~\ref{fig:late-fork-reward}. The spikes in the graph show the agent alternates between querying with probability zero (at the unavoidable action) and querying with positive probability (at the fork), just as the Bayes optimal agent does.\footnote{For ``Unknown Transitions'' the agent knows that actions are unavoidable while not knowing where they lead.} 
%We also found that if the query cost is set to zero, the agent almost always queries and if the cost is very high, the agent never queries. 
%While BAMCP++ does well, there is room for improvement. With the transition function known, performance is not better than random for 13 episodes. This is likely due to the MCTS not recognising the value of querying for short horizons. 

\subsubsection{BAMCP++ on Random MDPs}
BAMCP++ does well on very small MDPs like Bandits and 3-state Late Fork. Can it scale to larger and more varied MDPs? We compare BAMCP++, MCCH, and First-N on the Fork environments (Fig~\ref{fig:fork}) and on random MDPs with 5 states and 3 actions. The query cost is $c \seq 0.5$ throughout. To generate 25 random MDPs for testing algorithms, we sample rewards and transitions from symmetric Dirichlet distributions with $\alpha\seq 0.5$ and $\alpha \seq 0.2$ respectively. We call this the \textit{generating prior}. The BAMCP++ agent uses the generating prior across all MDPs (including the Fork environments) and uses a fixed number of MC simulations (200,000).

BAMCP++ and First-N use a fixed set of hyperparameters for all MDPs in Table~\ref{table:random}. These are set by gridsearch on random MDPs sampled from the generating prior. So the hyperparameters are tuned for the task ``Rand-25'' but not for any other tasks in Table~\ref{table:random}. We tried fixing hyperparameters for MCCH in the same way but performance was so poor that we instead tuned hyperparameters for each row in Table~\ref{table:random}.

\begin{figure}[t!]
\centering
\includegraphics[width=7.0cm]{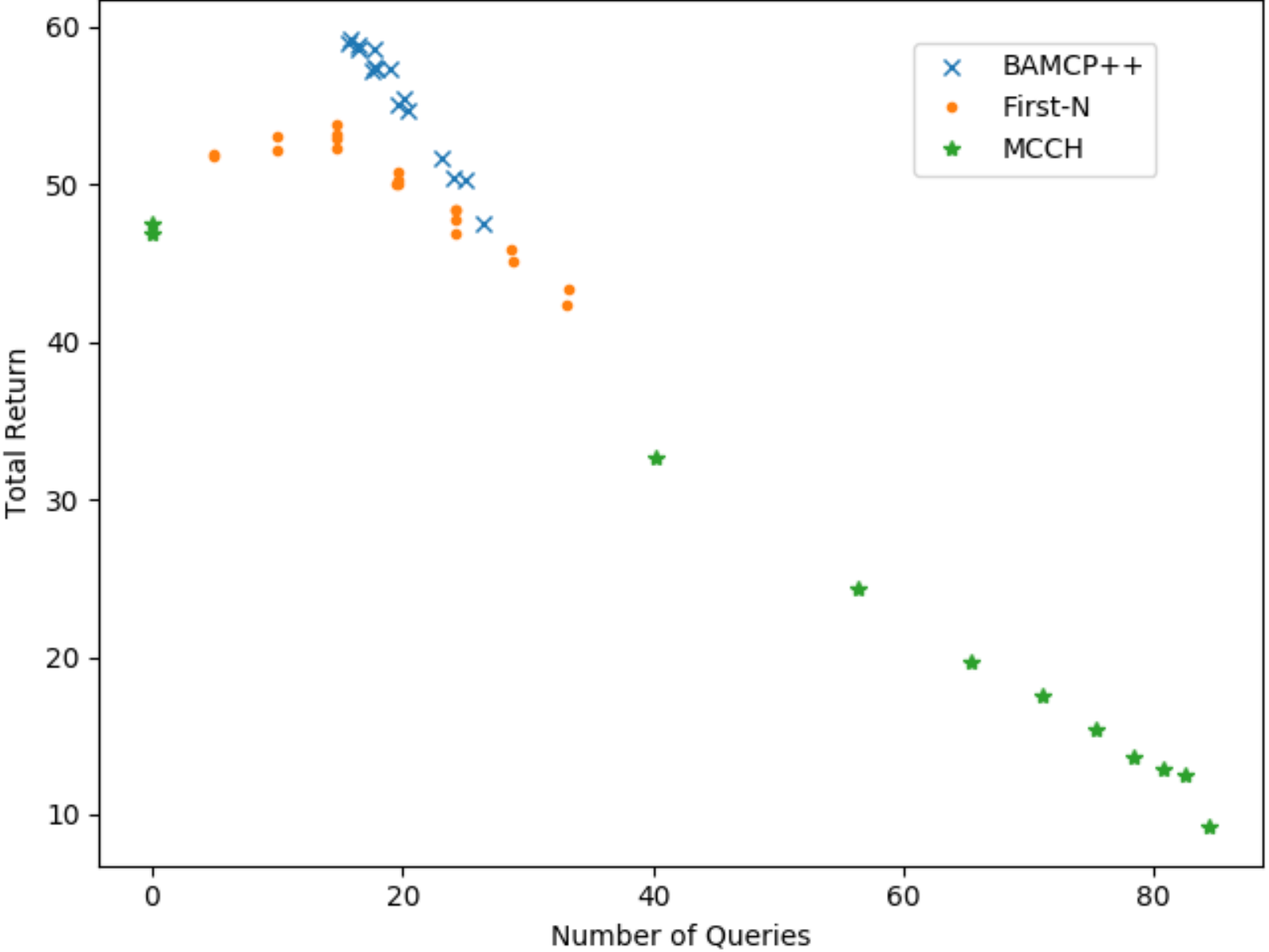}
\caption{Number of queries vs.\ total return on random MDPs for different algorithms and with varying hyperparameters.}
\label{fig:model-vs-heur}
\end{figure}

On random MDPs, BAMCP++ substantially outperforms the model-free approaches. The mean performance averaged over all 25 random MDPs is shown in row ``Rand-25'' of Table~\ref{table:random}. Here each algorithm has its hyperparameters tuned to the task. Figure~\ref{fig:model-vs-heur} shows performance (total return vs.\ number of queries) on the same task but with a range of different hyperparameter settings. MCCH performs poorly because without tuning of hyperparameters it queries far too much. First-N and BAMCP++ are both fairly robust to hyperparameter settings in terms of both number of queries and total return. BAMCP++ achieves more reward without querying more, suggesting it makes smarter choices of where to explore and which actions to query.
On Early and Late Fork environments, BAMCP++ performs best on horizon $T\seq30$; while First-N wins on horizon $T\seq50$. The Fork environments all have a maximum per-episode reward of 1 and hence a maximum total reward of 30 and 50 (for $T\seq30$ and $T\seq50$). As the horizon increases, BAMCP++ improves its absolute score significantly but its score declines as a function of the maximal total return. What explains this decline? The most challenging task ``Early5-50'' is a 10-state MDP with planning horizon of 250 timesteps (50 episodes $\times$ 5 steps per episode). This is a vastly larger search tree than for ``Late4-30'' but the number of MCTS simulations at each timestep was the same, making it harder to sample the best exploration strategies.

MCCH and First-N initially query all states indiscriminately. 
As the horizon increases, they scale scale well because there is more time for their indiscriminate querying to be exploited. The strong overall performance of First-N is partly due to our choice of MDPs. All reward distributions were Bernoulli (which have an upper-bound on their variance) and differences between optimal $Q^*(s,a)$ values for actions were rarely very small. So by tuning the hyperparameter $\mathrm{N}$ (the maximum number of queries per action) on random MDPs, First-N was well adapted to all the MDPs in Table~\ref{table:random}. But outside our experiments the same MDP could have reward distributions with huge variation in variance (e.g.\ Gaussian rewards with $\sigma^2\seq1$ and $\sigma^2\seq1000$) and so a Bayes optimal ARL agent would need to query some actions many more times than others.

\begin{table}[H]
\begin{center}
\begin{tabular}{l l l l} 
 & \textbf{BAMCP++} & \textbf{MCCH} & \textbf{First-N}\\\hline
Rand-25 & \textbf{60.2[8.8]} & 48.5[17] & 55.7[16]\\\hline
Late4-30 & \textbf{28.2[0.7]} & 25.1[9.0] & 26.1[2.9] \\
Late5-30 & \textbf{27.4[0.2]} & 25.7[8.2] & 25.6[2.5] \\
Late4-50 & 45.2[0.7] & 41.7[12] & \textbf{46.3[7.6]} \\
Late5-50 & 43.5[1.0] & 42.1[13] & \textbf{45.2[5.3] }\\\hline
Early4-30 & \textbf{25.9[1.9]} & 22.8[11] & 24.5[2.5]\\
Early5-30 & \textbf{23.8[3.7]} & 22.8[10] & 23.7[2.5]\\
Early4-50 & 41.2[3.2] & 40.1[17] & \textbf{43.2[8.8]}\\
Early5-50 & 32.9[6.9]& 39.3[16] & \textbf{42.9[5.5]}\\
%Late3-30 & 28.3[0.6] & 27.05[1.34] & 28.01[1.23]\\
%Late3-40 & 37.67[0.7] & 34.42[8.8] & 36.26[2.51] \\
%Late3-50 & 46.7[0.63] & 43.26[9.69] & 45.68[9.3] \\
%Late4-30 & 28.2[0.7] & 25.16[8.98] & 26.08[2.94] \\
%Late4-40 & 36.9[0.55] & 32.2[11.95] & 35.78[1.94] \\
%Late4-50 & 45.17[0.67] & 41.7[12.15] & 46.32[7.6] \\
%Late5-30 & 27.37[0.22] & 25.7[8.20] & 25.64[2.52] \\
%Late5-40 & 35.53[0.62] & 33.84[10.48] & 35.32[3.33] \\
%Late5-50 & 43.47[0.96] & 42.14[12.48] & 45.24[5.32] \\
%Early3-30 & 26.97[2.7]  & 26.82[1.38] & 26.84[1.75] \\
%Early3-40 & 35.67[3.27] & 32.16[13.43] & 35.56[2.16]\\
%Early3-50 & 43.43[3.55] & 44.7[12.21] & 44.8[5.87]\\
%Early4-30 & 25.87[1.9] & 22.76[11.26] & 24.48[2.54]\\
%Early4-40 & 32.97[3.96] & 32.18[12.49] & 34.38[2.13]\\
%Early4-50 & 41.2[3.24] & 40.08[16.78] & 43.2[8.81]\\
%Early5-30 & 23.8[3.68] & 22.8[10.31] & 23.64[2.52]\\
%Early5-40 & 28.43[7.27] & 30.84[12.87] & 34.28[10.08]\\
%Early5-50 & 32.93[6.91]& 39.28[16.41] & 42.88[5.48]\\
\end{tabular}
\end{center}
\caption{Mean (SD) of returns for different MDPs. ``Rand-25'' is the mean score over 25 random MDPs. ``Late4-30'' is the average over multiple runs on the Late Fork MDP with $N\seq4$ and horizon $T\seq30$. ``Early4-30'' is the corresponding average for the Early Fork MDP. Hyperparameters are fixed for BAMCP++ and First-N but tuned to each class of MDPs for MCCH.}
\label{table:random}
\end{table}

\section{CONCLUSION}

Active RL is a twist on standard RL in which the cost of evaluating the reward of actions is incorporated into the agent's objective. It is motivated by settings where rewards are constructed incrementally online, as when humans provide feedback to a learning agent. We introduced BAMCP++, an algorithm for Bayesian ARL in tabular MDPs which converges to the Bayes optimal policy in the limit of Monte-Carlo samples. In experiments, BAMCP++ achieves near-optimal performance on small MDPs and outperforms model-free algorithms on MDPs with 15 actions and a horizon of 100 timesteps. 

The key idea behind BAMCP++ is that MCTS is guided by a sophisticated (and more computationally costly) model-free learner in the rollouts. This helps alleviate a fundamental challenge for simulation-based ARL algorithms. Such algorithms must simulate recouping the upfront query costs by exploiting the information gained from queries. This requires simulations that are non-random (to capture exploitation) over many timesteps (query costs are only recouped after many timesteps).

%Active RL is ripe for future work. In tabular MDPs, function approximation \citep{BAMCPthesis} could be applied to improve sharing between similar BAMDP hyperstates. Flexible, LSTM-based meta-learning approaches could be adapted from RL \citep{LSTM} to ARL. ARL could also be explored in the setting of large discrete or continuous state spaces. In many such environments, model-based approaches based on MCTS are not competitive but the model-free heuristics here could provide a basic starting point. 

%Active RL also has practical applications. If rewards are constructed incrementally during learning and if their cost is significant relative to the agent's long-term earnings, then ARL is potentially applicable. One class of examples is where RL rewards are based on human responses to some input. This applies to medical trials where evaluating patient response to a treatment requires a doctor's assessment of the patient. This also applies to in-depth A-B testing, where users provide detailed feedback on the versions of the product being tested. 

%%%%% HIDE FOR ANON SUBMISSION
\newpage
\section*{Acknowledgements}
OE was supported by the Future of Humanity Institute (University of Oxford) and the Future of Life Institute grant 2015-144846. SSch is in a PhD position supported by Dyson. Clare Lyle contributed to early work on model-free heuristics and suggested the Early Fork environment. We thank Joelle Pineau and Jan Leike for helpful conversations. We thank David Abel, Michael Osborne and Thomas McGrath for comments on a draft. 

\bibliography{references}

\begin{thebibliography}{34}
\providecommand{\natexlab}[1]{#1}
\providecommand{\url}[1]{\texttt{#1}}
\expandafter\ifx\csname urlstyle\endcsname\relax
  \providecommand{\doi}[1]{doi: #1}\else
  \providecommand{\doi}{doi: \begingroup \urlstyle{rm}\Url}\fi

\bibitem[Abbeel and Ng(2004)]{apprentice}
P.~Abbeel and A.~Y. Ng.
\newblock Apprenticeship learning via inverse reinforcement learning.
\newblock In \emph{Proceedings of the twenty-first international conference on
  Machine learning}, 2004.

\bibitem[Anthony et~al.(2017)Anthony, Tian, and Barber]{slowandfast}
T.~Anthony, Z.~Tian, and D.~Barber.
\newblock Thinking fast and slow with deep learning and tree search.
\newblock In \emph{Advances in Neural Information Processing Systems}, pages
  5366--5376, 2017.

\bibitem[Araya et~al.(2012)Araya, Buffet, and Thomas]{BOLT}
M.~Araya, O.~Buffet, and V.~Thomas.
\newblock Near-optimal brl using optimistic local transitions.
\newblock \emph{arXiv preprint arXiv:1206.4613}, 2012.

\bibitem[Auer et~al.(2002)Auer, Cesa-Bianchi, and Fischer]{UCB1}
P.~Auer, N.~Cesa-Bianchi, and P.~Fischer.
\newblock Finite-time analysis of the multiarmed bandit problem.
\newblock \emph{Machine learning}, 47\penalty0 (2-3):\penalty0 235--256, 2002.

\bibitem[Azar et~al.(2017)Azar, Osband, and Munos]{azar2017minimax}
M.~G. Azar, I.~Osband, and R.~Munos.
\newblock Minimax regret bounds for reinforcement learning.
\newblock \emph{arXiv preprint arXiv:1703.05449}, 2017.

\bibitem[Bubeck et~al.(2012)Bubeck, Cesa-Bianchi, et~al.]{bubeck2012regret}
S.~Bubeck, N.~Cesa-Bianchi, et~al.
\newblock Regret analysis of stochastic and nonstochastic multi-armed bandit
  problems.
\newblock \emph{Foundations and Trends{\textregistered} in Machine Learning},
  5\penalty0 (1):\penalty0 1--122, 2012.

\bibitem[Castronovo et~al.(2016)Castronovo, Ernst, Cou{\"e}toux, and
  Fonteneau]{BenchBRL}
M.~Castronovo, D.~Ernst, A.~Cou{\"e}toux, and R.~Fonteneau.
\newblock Benchmarking for bayesian reinforcement learning.
\newblock \emph{PloS one}, 11\penalty0 (6), 2016.

\bibitem[Christiano et~al.(2017)Christiano, Leike, Brown, Martic, Legg, and
  Amodei]{HumanPref}
P.~F. Christiano, J.~Leike, T.~Brown, M.~Martic, S.~Legg, and D.~Amodei.
\newblock Deep reinforcement learning from human preferences.
\newblock In \emph{Advances in Neural Information Processing Systems}, pages
  4302--4310, 2017.

\bibitem[D.~Sadigh et~al.(2017)D.~Sadigh, S., and S.]{dorsa2017active}
A.~D.~Dragan D.~Sadigh, Shankar S., and Sanjit~A S.
\newblock Active preference-based learning of reward functions.
\newblock In \emph{Robotics: Science and Systems (RSS)}, 2017.

\bibitem[Daniel et~al.(2014)Daniel, Viering, Metz, Kroemer, and
  Peters]{ActiveRewards}
C.~Daniel, M.~Viering, J.~Metz, O.~Kroemer, and J.~Peters.
\newblock Active reward learning.
\newblock In \emph{Robotics: Science and System}, 2014.

\bibitem[Dearden et~al.(1998)Dearden, Friedman, and Russell]{BayesianQ}
R.~Dearden, N.~Friedman, and S.~Russell.
\newblock Bayesian q-learning.
\newblock In \emph{AAAI}, pages 761--768. AAAI Press, 1998.

\bibitem[Dragan(2017)]{dragan2017robot}
A.~D. Dragan.
\newblock Robot planning with mathematical models of human state and action.
\newblock \emph{arXiv preprint arXiv:1705.04226}, 2017.

\bibitem[Epshteyn et~al.(2008)Epshteyn, Vogel, and DeJong]{epshteyn2008active}
A.~Epshteyn, A.~Vogel, and G.~DeJong.
\newblock Active reinforcement learning.
\newblock In \emph{Proceedings of the 25th international conference on Machine
  learning}, pages 296--303. ACM, 2008.

\bibitem[Evans et~al.(2016)Evans, Stuhlm{\"u}ller, and
  Goodman]{evans2016learning}
O.~Evans, A.~Stuhlm{\"u}ller, and N.~D. Goodman.
\newblock Learning the preferences of ignorant, inconsistent agents.
\newblock In \emph{Proceedings of the Thirtieth AAAI Conference on Artificial
  Intelligence}, pages 323--329. AAAI Press, 2016.

\bibitem[Ghavamzadeh et~al.(2015)Ghavamzadeh, Mannor, Pineau, Tamar,
  et~al.]{BRLsurvey}
M.~Ghavamzadeh, S.~Mannor, J.~Pineau, A.~Tamar, et~al.
\newblock Bayesian reinforcement learning: A survey.
\newblock \emph{Foundations and Trends{\textregistered} in Machine Learning},
  8\penalty0 (5-6):\penalty0 359--483, 2015.

\bibitem[Guez(2015)]{BAMCPthesis}
A.~Guez.
\newblock \emph{Sample-Based Search Methods For Bayes-Adaptive Planning}.
\newblock PhD thesis, UCL (University College London), 2015.

\bibitem[Guez et~al.(2012)Guez, Silver, and Dayan]{BAMCP}
A.~Guez, D.~Silver, and P.~Dayan.
\newblock Efficient bayes-adaptive reinforcement learning using sample-based
  search.
\newblock In \emph{Advances in Neural Information Processing Systems}, pages
  1025--1033, 2012.

\bibitem[Ho and Ermon(2016)]{GAIL}
J.~Ho and S.~Ermon.
\newblock Generative adversarial imitation learning.
\newblock In \emph{Advances in Neural Information Processing Systems}, pages
  4565--4573, 2016.

\bibitem[Judah et~al.(2012)Judah, Fern, and Dietterich]{judah2012active}
K.~Judah, A.~Fern, and T.~G. Dietterich.
\newblock Active imitation learning via reduction to iid active learning.
\newblock In \emph{UAI}, pages 428--437, 2012.

\bibitem[Kolter and Ng(2009)]{BEB}
J.~Z. Kolter and A.~Y Ng.
\newblock Near-bayesian exploration in polynomial time.
\newblock In \emph{Proceedings of the 26th Annual International Conference on
  Machine Learning}, pages 513--520. ACM, 2009.

\bibitem[Krueger et~al.(2016)Krueger, Leike, Evans, and Salvatier]{activeRL}
D.~Krueger, J.~Leike, O.~Evans, and J.~Salvatier.
\newblock Active reinforcement learning: Observing rewards at a cost.
\newblock In \emph{Future of Interactive Learning Machines, NIPS Workshop},
  2016.

\bibitem[Kuleshov and Precup(2014)]{medicalBandits}
V.~Kuleshov and D.~Precup.
\newblock Algorithms for multi-armed bandit problems.
\newblock \emph{arXiv preprint arXiv:1402.6028}, 2014.

\bibitem[Osband et~al.(2013)Osband, Russo, and Van~Roy]{PSRL}
I.~Osband, D.~Russo, and B.~Van~Roy.
\newblock (more) efficient reinforcement learning via posterior sampling.
\newblock In \emph{Advances in Neural Information Processing Systems}, pages
  3003--3011, 2013.

\bibitem[Saunders et~al.(2017)Saunders, Sastry, Stuhlmueller, and
  Evans]{saunders2017trial}
W.~Saunders, G.~Sastry, A.~Stuhlmueller, and O.~Evans.
\newblock Trial without error: Towards safe reinforcement learning via human
  intervention.
\newblock \emph{arXiv preprint arXiv:1707.05173}, 2017.

\bibitem[Scott(2015)]{scott2015multi}
S.~L. Scott.
\newblock Multi-armed bandit experiments in the online service economy.
\newblock \emph{Applied Stochastic Models in Business and Industry},
  31\penalty0 (1):\penalty0 37--45, 2015.

\bibitem[Settles(2012)]{activeLearning}
B.~Settles.
\newblock Active learning.
\newblock \emph{Synthesis Lectures on Artificial Intelligence and Machine
  Learning}, 6:\penalty0 1--114, 2012.

\bibitem[Silver et~al.(2016)Silver, Huang, Maddison, Guez, Sifre, Van
  Den~Driessche, et~al.]{alphago}
D.~Silver, A.~Huang, C.~J. Maddison, A.~Guez, L.~Sifre, G.~Van Den~Driessche,
  et~al.
\newblock Mastering the game of go with deep neural networks and tree search.
\newblock \emph{Nature}, 529\penalty0 (7587):\penalty0 484--489, 2016.

\bibitem[Silver et~al.(2017)Silver, Schrittwieser, Simonyan, Antonoglou, Huang,
  Guez, et~al.]{alphazero}
D.~Silver, J.~Schrittwieser, K.~Simonyan, I.~Antonoglou, A.~Huang, A.~Guez,
  et~al.
\newblock Mastering the game of go without human knowledge.
\newblock \emph{Nature}, 550\penalty0 (7676):\penalty0 354, 2017.

\bibitem[Srinivas et~al.(2009)Srinivas, Krause, Kakade, and
  Seeger]{srinivas2009gaussian}
N.~Srinivas, A.~Krause, S.~M. Kakade, and M.~Seeger.
\newblock Gaussian process optimization in the bandit setting: No regret and
  experimental design.
\newblock \emph{arXiv preprint arXiv:0912.3995}, 2009.

\bibitem[Strens(2000)]{strens}
M.~Strens.
\newblock A bayesian framework for reinforcement learning.
\newblock In \emph{Proceedings of the 17th international conference on Machine
  learning}, 2000.

\bibitem[Subramanian et~al.(2016)Subramanian, Isbell~Jr, and
  Thomaz]{subramanian2016exploration}
K.~Subramanian, C.~L. Isbell~Jr, and Andrea~L. Thomaz.
\newblock Exploration from demonstration for interactive reinforcement
  learning.
\newblock In \emph{Proceedings of the 2016 International Conference on
  Autonomous Agents \& Multiagent Systems}, pages 447--456. International
  Foundation for Autonomous Agents and Multiagent Systems, 2016.

\bibitem[Sutton and Barto(1998)]{suttonBarto}
R.~S. Sutton and A.~G. Barto.
\newblock \emph{Reinforcement learning: An introduction}, volume~1.
\newblock MIT press Cambridge, 1998.

\bibitem[Warnell et~al.(2017)Warnell, Waytowich, Lawhern, and Stone]{DeepTAMER}
G.~Warnell, N.~Waytowich, V.~Lawhern, and P.~Stone.
\newblock Deep tamer: Interactive agent shaping in high-dimensional state
  spaces.
\newblock \emph{arXiv preprint arXiv:1709.10163}, 2017.

\bibitem[Wirth et~al.(2017)Wirth, Akrour, Neumann, F{\"u}rnkranz,
  et~al.]{wirth2017survey}
C.~Wirth, R.~Akrour, G.~Neumann, J.~F{\"u}rnkranz, et~al.
\newblock A survey of preference-based reinforcement learning methods.
\newblock \emph{Journal of Machine Learning Research}, 18\penalty0
  (136):\penalty0 1--46, 2017.

\end{thebibliography}

\end{document}